%% file: main.tex
\documentclass[lettersize,journal]{IEEEtran}
\usepackage{amsmath,amsfonts}
\usepackage{algorithmic}
\usepackage{algorithm}
\usepackage{array}
\usepackage[caption=false,font=normalsize,labelfont=sf,textfont=sf]{subfig}
\usepackage{textcomp}
\usepackage{stfloats}
\usepackage{url}
\usepackage{verbatim}
\usepackage{graphicx}
\usepackage{cite}
\usepackage{booktabs}
\usepackage{enumitem}
\usepackage[dvipsnames]{xcolor}
\usepackage[most]{tcolorbox}
\usepackage{color,xcolor}

\begin{document}

\title{RVISA: Reasoning and Verification for Implicit Sentiment Analysis}


\author{Wenna Lai, Haoran Xie, Guandong Xu, Qing Li
\thanks{The research was supported by the Faculty Research Grant (DB24A4) at Lingnan University, Hong Kong. \emph{(Corresponding author: Haoran Xie.)}\\
\indent Wenna Lai is with the Department of Computing, Hong Kong Polytechnic University, Hong Kong, also the School of Computer Science and the Data Science Institute, University of Technology Sydney, Australia.\\
\indent Haoran Xie is with the School of Data Science, Lingnan University, Hong Kong (email: hrxie@ieee.org).\\
\indent Guandong Xu is with the School of Computer Science and the Data Science Institute, University of Technology Sydney, Sydney, NSW 2007, and also the Education University of Hong Kong, Hong Kong SAR.\\
\indent Qing Li is with the Department of Computing, Hong Kong Polytechnic University, Hong Kong SAR.
}}



\maketitle

\begin{abstract}
With an increasing social demand for fine-grained sentiment analysis (SA), implicit sentiment analysis (ISA) poses a significant challenge with the absence of salient cue words in expressions. It necessitates reliable reasoning to understand how the sentiment is aroused and thus determine implicit sentiments. In the era of Large Language Models (LLMs), Encoder-Decoder (ED) LLMs have gained popularity to serve as backbone models for SA applications, considering impressive text comprehension and reasoning ability among diverse tasks. On the other hand, Decoder-only (DO) LLMs exhibit superior natural language generation and in-context learning capabilities. However, their responses may contain misleading or inaccurate information. To identify implicit sentiment with reliable reasoning, this study proposes RVISA, a two-stage reasoning framework that harnesses the generation ability of DO LLMs and the reasoning ability of ED LLMs to train an enhanced reasoner. Specifically, we adopt three-hop reasoning prompting to explicitly furnish sentiment elements as cues. The generated rationales are utilized to fine-tune an ED LLM into a skilled reasoner. Additionally, we develop a straightforward yet effective verification mechanism to ensure the reliability of the reasoning learning. We evaluated the proposed method on two benchmark datasets and achieved state-of-the-art results in ISA performance.
\end{abstract}

\begin{IEEEkeywords}
Implicit sentiment analysis, Large language models, Multi-task learning, Chain-of-Thought.
\end{IEEEkeywords}

\input{src/introduction}
\input{src/related_work}
\input{src/methodology}
\input{src/experiment}
\input{src/discussion}
\input{src/conclusion}

{\appendix \input{src/Appendix}}
 
%

\bibliography{IEEEabrv, src/custom}

\vspace{-20pt} 


\begin{IEEEbiography}[{\includegraphics[width=1in, height=1.25in,clip,keepaspectratio]{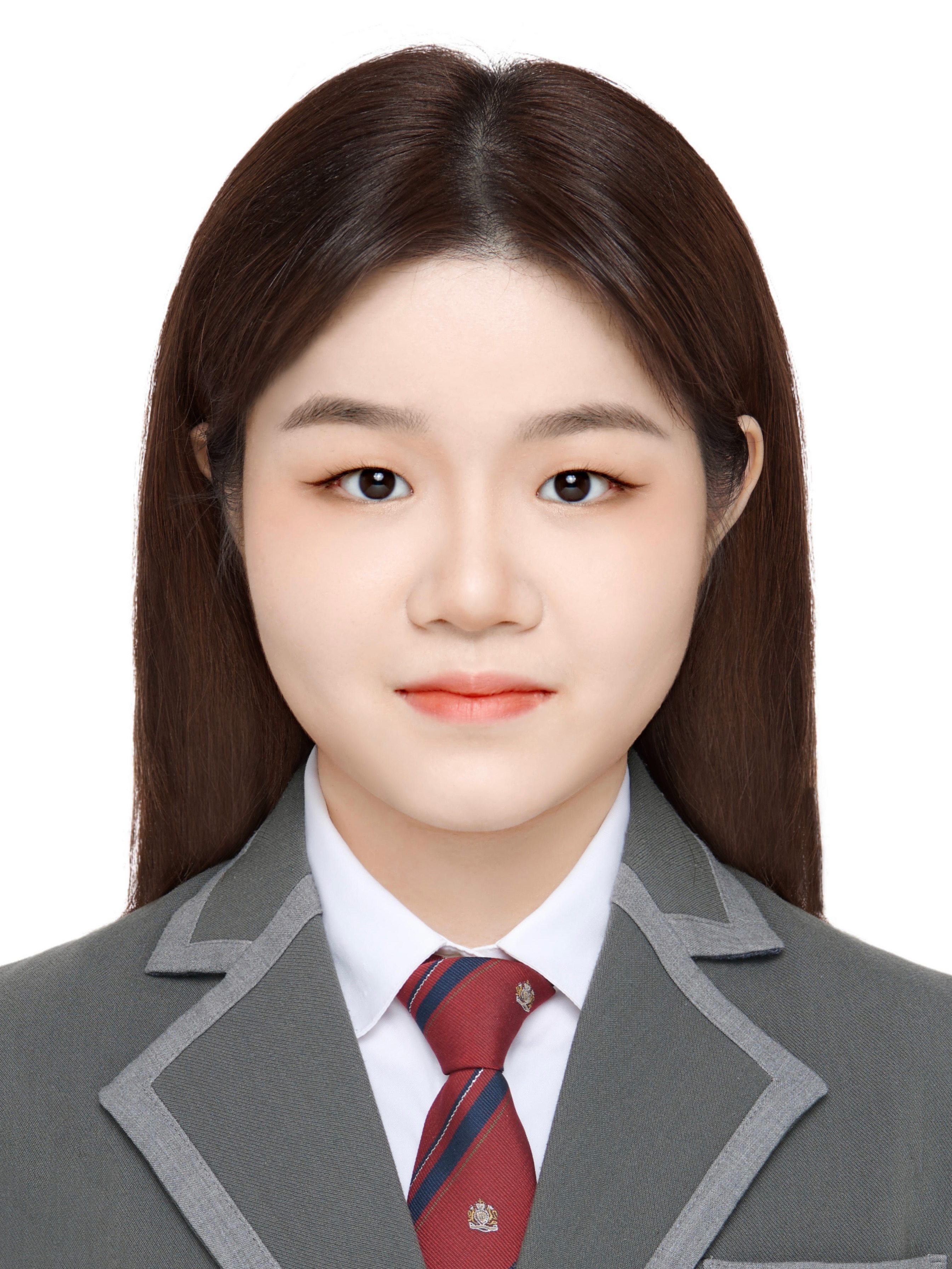}}]{Wenna Lai}
is currently a joint Ph.D. student at the Department of Computing, Hong Kong Polytechnic University, under the supervision of Prof. Qing Li, and the School of Computer Science, University of Technology Sydney, under the supervision of Prof. Guandong Xu. She has been working closely with Prof. Haoran Xie at the School of Data Science, Lingnan University, Hong Kong. Before that, she received her Master's degree in the Department of Electrical and Computer Engineering from the National University of Singapore. Her research interests include Affective Computing and NLP for Social Good.
\end{IEEEbiography}
\vspace{-25pt}
\begin{IEEEbiography}[{\includegraphics[width=1in,height=1.25in,clip,keepaspectratio]{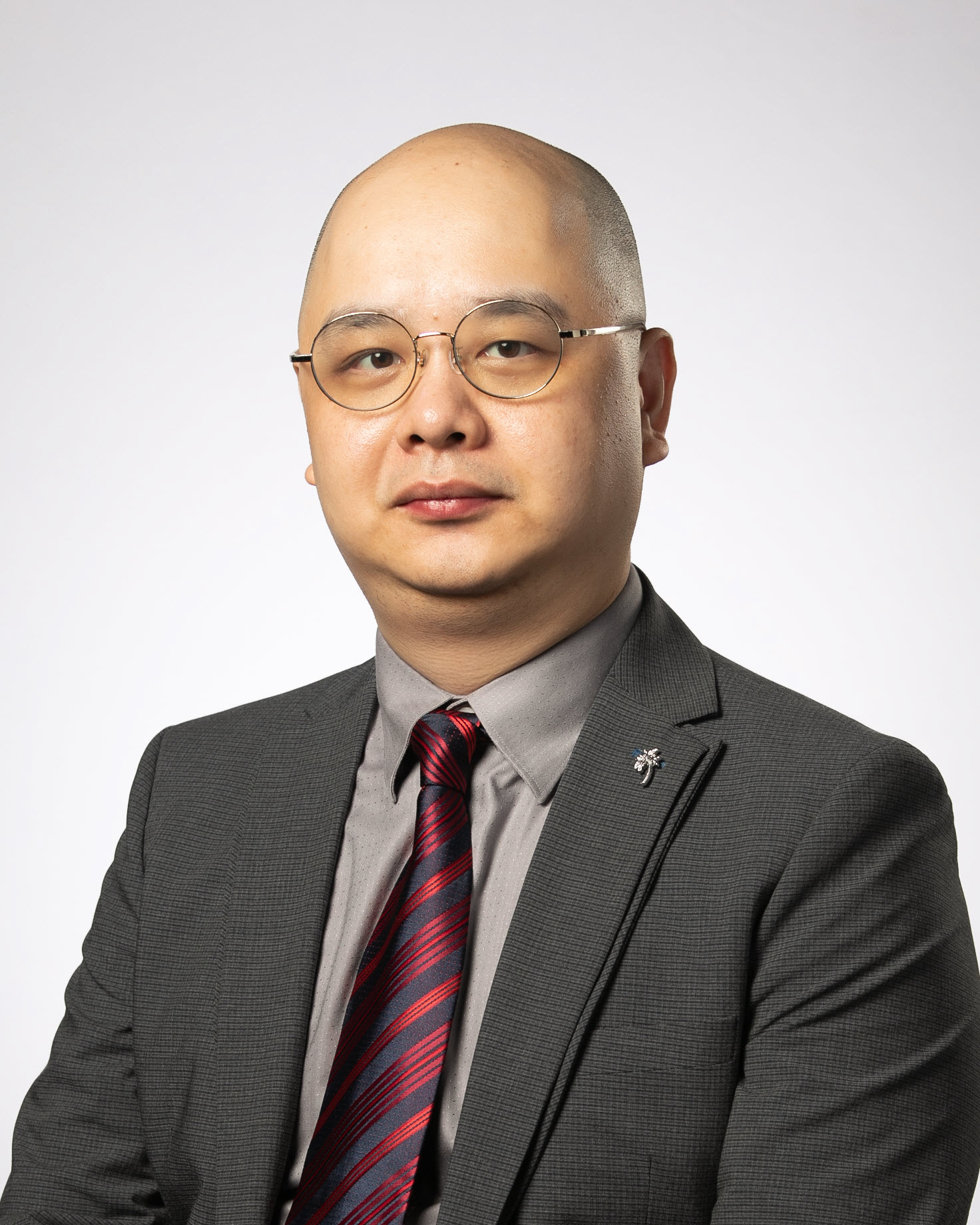}}]{Haoran Xie} (Senior Member, IEEE)
received a Ph.D. degree in Computer Science from City University of Hong Kong and an Ed.D degree in Digital Learning from the University of Bristol. He is currently the Acting Associate Dean and Associate Professor at the School of Data Science, Lingnan University, Hong Kong. His research interests include artificial intelligence, big data, and educational technology. He has published 411 research publications, including 236 journal articles such as IEEE TPAMI, IEEE TKDE, IEEE TAFFC, and IEEE TCVST. He is the Editor-in-Chief of Natural Language Processing Journal, Computers \& Education: Artificial Intelligence and Computers \& Education: X Reality. He has been selected as the World's Top 2\% Scientists by Stanford University.
\end{IEEEbiography}
\vspace{-25pt}
\begin{IEEEbiography}[{\includegraphics[width=1in,height=1.25in,clip,keepaspectratio]{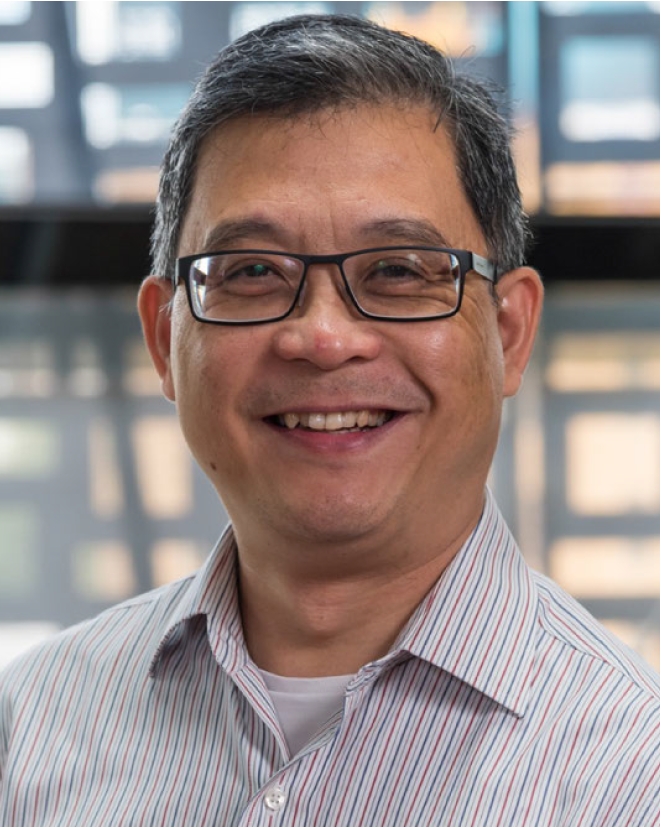}}]{Guandong Xu} (Member, IEEE) received the Ph.D. degree in computer science from Victoria University, Melbourne, VIC, Australia, in 2009. He is currently a Professor and a Program Leader at the School of Computer Science and Data Science Institute, University of Technology Sydney, Sydney, NSW, Australia. His research interests include data science, data analytics, recommender systems, web mining, user modeling, NLP, social network analysis, and social media mining.
\end{IEEEbiography}
\vspace{-25pt}
\begin{IEEEbiography}[{\includegraphics[width=1in,clip,keepaspectratio]{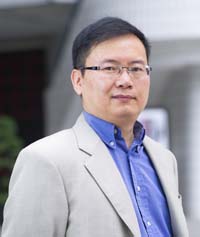}}]{Qing Li} (Fellow, IEEE) received the B.Eng. degree in Computer Science from Hunan Univeristy, Hunan, China, in 1982, and the M.S. and Ph.D. degrees in Computer Science from the University of Southern California, LA, California, USA, in 1985 and 1988, respectively. Qing Li is a Chair Professor and Head at the Department of Computing, The Hong Kong Polytechnic University. His research focuses on data science, web mining, and artificial intelligence.He is a Fellow of IET, a Fellow of IEEE, a member of ACM SIGMOD and IEEE Technical Committee on Data Engineering. He is the chairperson of the Hong Kong Web Society, and is a steering committee member of DASFAA, ICWL, and WISE Society.
\end{IEEEbiography}

\vfill

\end{document}

%% file: src/introduction.tex
\section{Introduction}

\IEEEPARstart{S}{\lowercase{entiment}} analysis (SA) aims to evoke opinions, sentiments, and emotions through different computational methods\cite{Liu_2012}. Nowadays, people have demonstrated a stronger willingness to express and share their ideas online about day-to-day activities and global issues. With the increasing demand on social media, SA has gained significant interest considering great commercial value in exploring customer opinions or sentiments from user reviews or other sources of information. Meanwhile, sentiments can assist learning, communication, decision-making, and situation awareness in human-centric environments \cite{Das2023MultimodalSA}.
\begin{figure}[h]
\includegraphics[width=0.5\textwidth]{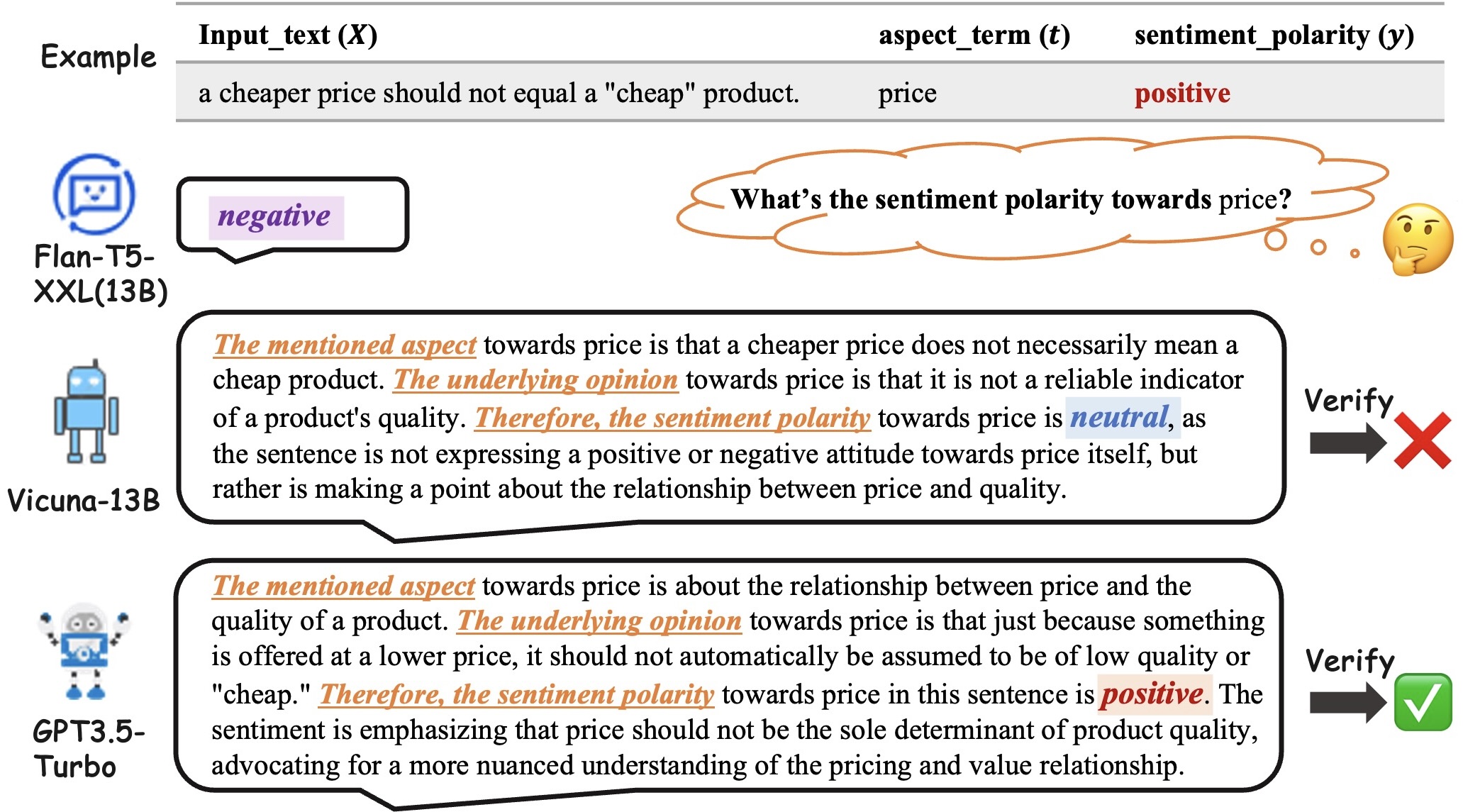}
\caption{Different LLMs demonstrate \textbf{diverse reasoning abilities} for implicit sentiment analysis. It is conducive to detecting implicit sentiment polarities by explicitly inferring sentiment elements as rationale but \textbf{verification is required to ensure reliability}.}
\label{fig:example}
\end{figure}
Traditionally, SA is classified into three levels, which are document-level, sentence-level, and aspect-level \cite{Cui_Wang_Ho_Cambria_2023}. While the document and sentence level analyze the sentiment towards the overview of a document or sentence, aspect-based sentiment analysis (ABSA) is more fine-grained to extract the opinion towards a given aspect or entity. In many cases, there may be multiple aspects in one sentence, making it challenging to pinpoint a specific target and identify the corresponding sentiment. 

Considering context information, sentiment analysis can be further classified into implicit sentiment analysis (ISA) and explicit sentiment analysis (ESA), where expressions in ISA contain no explicit polarity markers but still deliver human-aware sentiment polarity \cite{Russo_Caselli_Strapparava_2015}. In 2021, \cite{Li2021LearningIS} split the SemEval-2014 Restaurant and Laptop benchmarks into Explicit Sentiment Expression slice and Implicit Sentiment Expression slice based on the presence of opinion words, drawing attention to ISA in ABSA tasks. \cite{Fei_Li_Liu_Bing_Li_Chua} conducted pre-experiments on 20 existing sentiment classifiers and investigated that traditional methods performed ineffectively towards the same implicit case. They suggested that majority of traditional classifiers tend to overlook ISA problem and address ISA superficially. Although humans can easily grasp real intent and perceive changes in mood with common sense and reasoning ability, it is more difficult for models to tackle ISA than ESA, due to limited context information and insufficient reasoning skills. 

As recent great triumph of large language models (LLMs) has demonstrated impressive complex reasoning with chain-of-thought (CoT) prompting \cite{Wei2022ChainOT, Kojima_Shixiang_Gu_Reid_Matsuo_Iwasawa} and in-context learning ability \cite{Wei2021FinetunedLM}, more scholars tend to embrace LLMs for downstream applications \cite{HoSY23, Rajani_McCann_Xiong_Socher_2019, HsiehLYNFRKLP23}. \cite{Zhang2023SentimentAI} investigated the performance of LLMs in prompt-based inference and observed that for tasks requiring structured sentiment output, like ABSA tasks, both DO LLMs (e,g., GPT-3.5-turbo \cite{CHATGPT}) and ED LLMs (e.g., Flan-T5-XXL \cite{flant5}) tend to lag behind ED backbone models (e.g., T5-Large \cite{Raffel2019ExploringTL}) trained with domain-specific data in automatic and human evaluations. The performance can vary significantly with different prompt designs. These indicate that deploying LLMs for ISA directly without training may not fully unleash their reasoning capacity for achieving satisfactory results. \cite{Fei_Li_Liu_Bing_Li_Chua} first employed CoT fine-tuning on Flan-T5 for ISA and gained improved performance. However, intermediate steps generated by Flan-T5 were most likely to be untrustworthy, with insufficient or duplicate content constrained by weak generation capacity.  As illustrated in Figure \ref{fig:example}, different LLMs performed diversely in analyzing implicit sentiment towards the aspect term `\emph{price}', given the text `\emph{a cheaper price should not equal a ``cheap'' product}'. Inferior models, like Flan-T5 in the group of Encoder-Decoder (ED) LLMs, displayed excellent comprehension and reasoning in solving tasks with diverse input information, but limited generation and prompt-based inference capabilities on open-text \cite{Tay00GW0CBSZZHM23}. They were predisposed to inaccurately predict implicit sentiment in the absence of explicit cues. Conversely, Decoder-only (DO) LLMs with more advanced generation ability, such as Vicuna-13B \cite{zheng2023judging} and GPT-3.5-turbo, demonstrated enhanced proficiency in explicitly deducing sentiment elements pertinent to the context under reasoning prompts, while reliability in achieving accurate or correct responses was not guaranteed. Moreover, LLMs often showcase superior performance with emergent abilities when scaling up at a certain level \cite{emergent22}, the direct deployment or fine-tuning of large-scale models (e.g., GPT-3.5-turbo) might be hindered by considerable computational costs. To effectively discern implicit sentiment polarities towards a specified aspect, it is essential to exploit reliable reasoning methods for applicable backbone models.

With this motivation, we attempt to equip ED backbone models with enhanced reasoning ability by explicitly learning from convincing rationales provided by DO LLMs through synchronous verification. Specifically, we follow the sentiment element construction and design corresponding three-hop reasoning prompting to guide DO LLMs in explicitly inferring sentiment elements before determining the final sentiment. Then an ED model is served as the backbone model and fine-tuned based on the generated rationales and golden labels in datasets. To ensure the quality of reasoning learning, we further introduce an answer-based verification mechanism as an additional signal to assess the reliability of the rationale, which promotes dialectical learning to identify and rectify potential inaccuracies.

In summary, the contributions of this work are as follows:
\begin{itemize}
    \item We propose a novel two-stage learning framework, Reasoning and Verification for Implicit Sentiment Analysis (RVISA), marking the endeavor to improve the proficiency of ED backbone models as adept reasoners in ISA, complemented by the generative strengths of DO LLMs. 
    \item We introduce a straightforward yet efficacious verification mechanism to provide reliable supervision for reasoning learning and improve overall performance.
    \item The evaluation outcomes on two benchmark datasets underscore the efficacy of our method in achieving state-of-the-art results in ISA performance.
\end{itemize} 

%% file: src/related_work.tex
\section{Related Work}
\begin{figure*}[h]
\includegraphics[width=\textwidth]{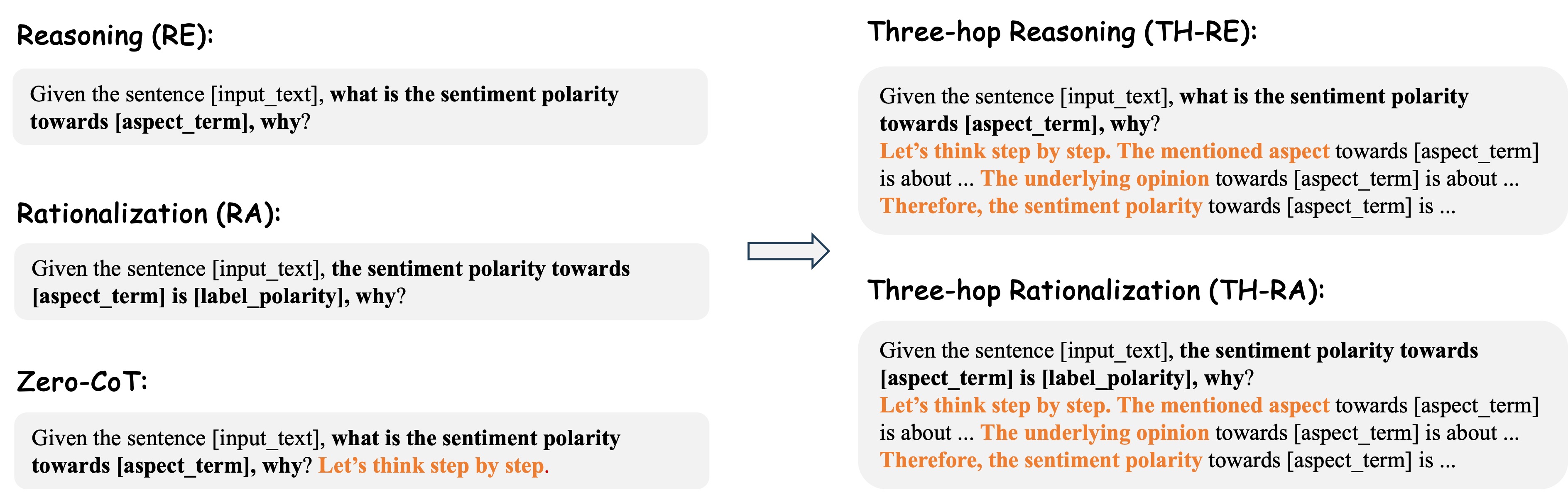}
\caption{Reasoning promptings applying to sentiment analysis. Left: commonly used prompting modes. Right: three-hop prompting for ISA.}
\label{fig:rationale}
\end{figure*}
In this work, we train a skilled reasoner with the cooperation of LLMs to conduct implicit sentiment analysis, learning fruitful information from rationales generated by reasoning prompting. We draw attention to the existing research on implicit sentiment analysis and methods that learn from reasoning prompting making use of emergent abilities showcased in LLMs.

\subsection{Implicit Sentiment Analysis}
Implicit sentiment analysis has gained considerable attention in the field of sentiment analysis \cite{Zong_Xia_Zhang_2021, Russo_Caselli_Strapparava_2015}. In the beginning, great efforts have been taken into solving the implicit sentiment detected in sentence level \cite{Xu_Wang_Feng_Yang_Zhang, Zhou_Wang_Zhang_He_2021}. With the increasing social demand, recent scholars attempted to develop effective paradigms tackling the unique characteristics of implicit sentiment analysis at a more fine-grained level towards the aspect target \cite{Li2021LearningIS, Wang_Zhou_Sun_Ye_Gui_Zhang_Huang_2022, Fei_Li_Liu_Bing_Li_Chua }.  
To capture the implicit sentiment expression, some research exploited extra knowledge to further improve the learning performance. \cite{Li2021LearningIS} pre-trained on large-scale sentiment annotated corpora with supervised contrastive learning objectives to align the representation of explicit and implicit sentiment expressions. Instead of making use of external knowledge, \cite{Ouyang_Yang_Liang_Wang_Wang_Li_2024} generated explicit sentiment augmentation based on the language model itself to enhance implicit classification tendencies. Considering the difficulties of obtaining the full knowledge through additional means, \cite{Wang_Zhou_Sun_Ye_Gui_Zhang_Huang_2022} proposed reasoning learning under causal intervention to capture the correlation within the expressions. The relationship within fine-grained sentiment analysis can be summarized into four key sentiment elements involving \emph{target, aspect, opinion, and sentiment polarity}, which are highly close to each other in understanding the underlying sentiment \cite{Peng_Xu_Bing_Huang_Lu_Si_2020}. With the impressive performance of chain-of-thought (CoT) and in-context learning abilities showcased in LLMs, \cite{Fei_Li_Liu_Bing_Li_Chua} introduced CoT fine-tuning to guide the ED backbone model inferring sentiment elements including implicit sentiment polarities step-by-step in an easy-to-hard manner. Similar to that, our approach makes use of fine-grained sentiment elements as cues for chain-of-thought prompting. But considering the limited generation capabilities of ED LLMs (e.g., Flan-T5 \cite{flant5}), rather than inferring the sentiment elements from backbone models themselves, we train ED backbone models to become proficient reasoners by leveraging the informative rationale generated from DO LLMs (e.g., GPT-3.5-turbo \cite{CHATGPT}).

\subsection{Reasoning Prompting}
LLMs have demonstrated impressive complex reasoning abilities with Chain-of-Thought (CoT) prompting \cite{Wei_Wang_Schuurmans_Bosma_Chi_Le_Zhou, Kojima_Shixiang_Gu_Reid_Matsuo_Iwasawa}. The use of reasoning prompting aims to guide the model in thinking step-by-step and leveraging most of the inference power for task solving. It is discovered effective in boosting the zero-shot or few-shot performance of LLMs \cite{Brown2020LanguageMA, Wang_Wei_Schuurmans_Le_Chi_Zhou, Fu_Peng_Sabharwal_Clark_Khot_2022, Zhang_Zhang_Li_Smola_2022}. Figure \ref{fig:rationale} illustrates various reasoning prompting applying to sentiment analysis. On the left-hand side are commonly used prompting modes including Reasoning and Rationalization:

\paragraph{Reasoning (RE)} \cite{hase-etal-2020-leakage}
introduced multi-task learning with reasoning prompting by simultaneously learning the question-answer pairs and question-explanation pairs. The generated rationales for the question will not have the ground truth answer for reference, which prompts the language model to infer the answer according to its step-by-step inference and own judgment. Therefore, the answer showcased in the explanation can be different from the gold label.
\paragraph{Rationalization (RA)}
\cite{Camburu} first proposed the idea of rationalization, which attempts to retrieve the explanation for the question by explicitly giving the correct answer. The intuition towards it is to rationalize the question with the golden label and provide the possible reasons behind the question-answer connection.

Besides them, \cite{Kojima_Shixiang_Gu_Reid_Matsuo_Iwasawa} revealed that LLMs are capable of incremental reasoning without exemplars. Simply by incorporating a prompt \emph{``let's think step by step''} (i.e., Zero-CoT in Figure \ref{fig:rationale}), it is universally applicable across tasks. However, the granularity of the reasoning steps generated by Zero-CoT remains unpredictable, hinging on the LLM's inherent knowledge and varying across models. Furthering this exploration, \cite{Jin_Yu_Shu_Zhao_Hua_Meng_Zhang_Du} examined the influence of reasoning step length within prompts and suggested that maintaining a certain step size according to the complexity of the task has a critical role in forming the final answer. 

Based on these insights, our method ingeniously builds on previous prompting methods by adding a three-hop strategy that uses the construction of sentiment elements to keep critical reasoning steps going, as shown on the right in Figure \ref{fig:rationale}. Considering the heuristic about answer inference, the explanation given by reasoning inference that leads to the correct answer should be more trustworthy for answer prediction. LLMs may falter in complex scenarios where reasoning prompts alone are insufficient, potentially yielding explanations riddled with inaccuracies. To mitigate this issue, \cite{Explanations} employed answer-based filtering to improve rationale quality, with Rationalization serving as the backup option for erroneous explanations. Rather than discarding inaccurate rationales outright or attempting to compensate for them with additional information sources, our approach retains these informative rationales throughout the learning process. 

\subsection{Learning from Rationale}
Learning from explanations and empowering the training model with reasoning abilities have been explored in various fields \cite{Rajani_McCann_Xiong_Socher_2019, Explanations, HsiehLYNFRKLP23, HoSY23}. LLMs are capable of validating their responses with reasonable intermediate steps 
\cite{Wei2022ChainOT, Kojima_Shixiang_Gu_Reid_Matsuo_Iwasawa}, rationales can be used as demonstrations \cite{Brown2020LanguageMA} or extra fine-tuning data \cite{HoSY23, Huang_Gu_Hou_Wu_Wang_Yu_Han_2022, Zelikman_Wu_Goodman} to improve the learning performance. Considering the training cost for LLMs, rationales can also serve as valuable supervised signals for smaller task-specific models, which can be more easily deployed \cite{Magister_Mallinson_Adamek_Malmi_Severyn_2022, Explanations, HsiehLYNFRKLP23, Rajani_McCann_Xiong_Socher_2019}. However, \cite{HsiehLYNFRKLP23} directly kept the answer generated by LLMs as supervision signals, which neglected the potential erroneous occurrence. \cite{Explanations} reorganized the rationale set based on answer-based filtering to mitigate the possibility of error learning, but potential misleading information may still exist even with the guidance of a correct answer. In contrast, our approach augments the overall performance by incorporating an answer-based verification mechanism as an additional layer of supervision for multi-task learning. This innovative strategy not only preserves valuable insights contained within the rationales but also leverages them to refine the learning process with both positive and negative signals. 

We compare various prompting methodologies and substantiate the superior performance of our three-hop prompting in the nuanced domain of implicit sentiment analysis through comprehensive experiments. Beyond that, the introduction of the verification mechanism further improves the performance. The integration of three-hop prompting with the verification mechanism effectively navigates complexities inherent in LLM reasoning process. 

%% file: src/methodology.tex
\section{Two-stage Reasoning Framework}
\begin{figure*}[h]
\includegraphics[width=\textwidth]{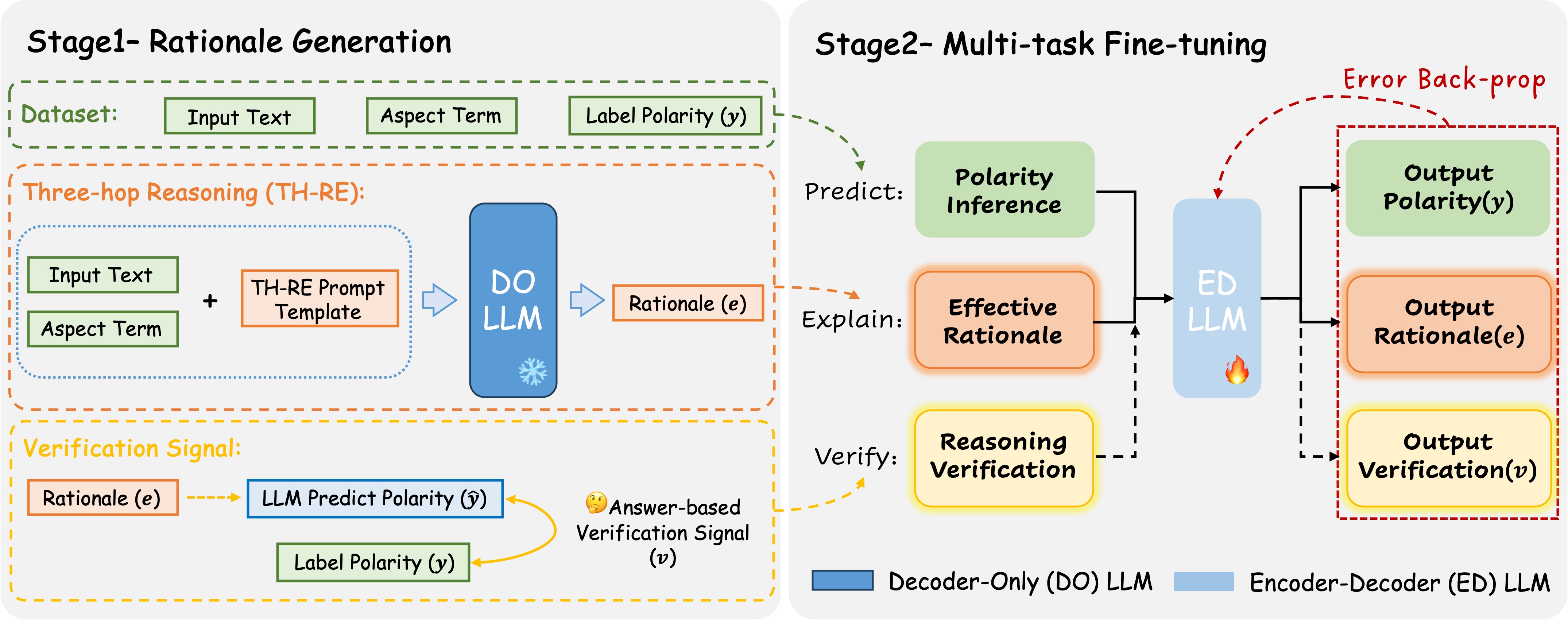}
\caption{The overview of proposed two-stage reasoning framework RVISA. Left: rationale generation stage leveraging DO LLM to prepare effective rationales and corresponding answer-based verification signals. Right: multi-task fine-tuning stage to train an ED backbone model as an enhanced reasoner with additional explanation tasks along with verification supervision.}
\label{fig:framework}
\end{figure*}
We propose a novel two-stage framework, RVISA (as shown in Figure \ref{fig:framework}), aiming to empower ED models with enhanced reasoning ability and incorporate the answer-based verification mechanism for reasoning refinement during model learning. In the initial stage, we leverage DO LLMs to generate insightful rationales and predict labels through our three-hop reasoning prompting approach. Then the verification signals are curated according to the correctness of LLM prediction labels. In the second stage, the generated rationales are employed for multi-task fine-tuning on an ED backbone model. To further ensure the reliability of the generated rationales, we implement a straightforward yet effective verification mechanism with an additional task supervised by the verification signals to guide self-revision during the reasoning learning process. Different tasks are distinguished by task-specific prefixes. It is thought that the model can learn to understand the underlying logic and relationships among sentiment elements that govern implicit sentiment prediction, by training on reasoning rationales and self-verification signals at the same time under the supervision of gold labels that have been annotated.
\subsection{Problem Definition} In sentiment analysis tasks including explicit sentiment analysis and implicit sentiment analysis, given the dataset $D = {(x_i, y_i)}^N$, where $1 \le i \le N$, $x_i$ represents an input sentence serving as a data example. Within each sentence $x_i$, an aspect term $t_i$ is identified, denoted as $t_i \subset x_i$. The relevant sentiment elements consist of aspect $a_i$, opinion $o_i$, and sentiment polarity $y_i$. The objective of the task is to infer the sentiment polarity $y_i$ towards the aspect term $t_i$, given the input sentence $x_i$ and the specified aspect term $t_i$. In the standard prompting approach for direct fine-tuning, the LLM predicts the sentiment polarity $\hat{y_i}$ solely via $\hat{y_i} = argmax p(y_i|x_i,t_i)$. Without considering the intermediate sentiment elements, this approach potentially limits the ability of models to capture the sentiment nuances present in the text.

\subsection{Three-hop Reasoning Generation}\label{generation}
To improve the generation of informative rationales, we prompt DO LLMs to generate intermediate steps during the inference of implicit sentiments. To understand how the sentiment is aroused, sentiment elements are essential in directing inference process since they contribute to constructing the complete picture of sentiment analysis. Therefore, we further design the three-hop prompting as illustrated in Figure \ref{fig:rationale}, deviating from conventional prompting modes. The objective of this design is to dominate reasoning process by extracting closely associated sentiment elements. Simultaneously, this design is conducive to standardizing the generative structure, facilitating improved learning of patterns and interconnections among rationales. The details of the three-hop reasoning prompting are explained as follows.

\paragraph{Three-hop Reasoning (TH-RE)}

Fine-grained sentiment analysis involves dissecting key sentiment elements involving the target, aspect, opinion, and sentiment polarity \cite{Peng_Xu_Bing_Huang_Lu_Si_2020}. Various approaches exist for solving these individual subtasks or their combinations, collectively contributing to a comprehensive sentiment analysis picture. To address the complexity of the task holistically, it is essential to consider the components systematically and tackle them incrementally. \cite{Fei_Li_Liu_Bing_Li_Chua} first design the prompting based on the CoT strategy by explicitly inferring the sentiment elements and then employ the prompting for three-step generation during model fine-tuning. However, the prompting for each step is inferred separately for a single sentiment element at a time, with the results concatenated as context information for the subsequent step. 

In our design, we adopt a structured approach by explicitly presenting sentiment elements in a natural language sequence to construct a three-hop reasoning prompt. This approach underscores the causal relationships among sentiment elements and the final sentiment polarity prediction in a single iteration.

As shown in the template below, we incorporate sentiment elements as cues at the end of \emph{``let's think step-by-step''}, guiding the language model to generate reasoning steps in alignment with the sentiment elements' understanding and finally infer the sentiment polarity. We expect DO LLMs to predict the explanation via $\hat{e_i} = argmax \space p(e_i|x_i,t_i)$, where $\hat{a_i}, \hat{o_i}, \hat{y_i} \subset \hat{e_i}$.

\begin{tcolorbox}[colback=gray!10,
                  colframe=black,
                  arc=1mm, auto outer arc,
                 ]
  Given the sentence $x_i$, what is the sentiment polarity towards $t_i$, why? 
Let’s think step by step. \newline The \textbf{mentioned aspect} towards $t_i$ is about ... The \textbf{underlying opinion} towards $t_i$ is about ... Therefore, the \textbf{sentiment polarity} towards $t_i$ is ...
\end{tcolorbox}

\paragraph{Three-hop Rationalization (TH-RE)}

Diverse from TH-RE, we integrate three-hop reasoning with rationalization to establish the Three-hop rationalization (TH-RA) prompting. Specifically, the gold label will be given as the reference, which prompts the LLMs to elucidate the annotated sentiment label through a systematic and step-by-step inference process guided by sentiment elements. We expect DO LLMs to predict the explanation via $\hat{e_i} = argmax \space p(e_i|y_i, x_i,t_i)$, where $\hat{a_i}, \hat{o_i}, \hat{y_i} \subset \hat{e_i}$.
\begin{tcolorbox}[colback=gray!10,
                  colframe=black,
                  arc=1mm, auto outer arc,
                 ]
 Given the sentence $x_i$, the sentiment polarity towards $t_i$ is $y_i$, why? 
Let’s think step by step. \newline The \textbf{mentioned aspect} towards $t_i$ is about ... The \textbf{underlying opinion} towards $t_i$ is about ... Therefore, the \textbf{sentiment polarity} towards $t_i$ is ...
\end{tcolorbox}

\subsection{Multi-task Fine-tuning}

We employ a multi-task fine-tuning approach to simultaneously learn the rationales generated by the LLM and the annotated labels. Given the dataset $D = \{(x_i, y_i)\}^N$, where $1 \le i \le N$, we generate an explanation $e_i$ to serve as a rationale for each input $x_i$ as detailed in Section \ref{generation}. Each explanation $e_i$ encompasses a generated label $\hat{y_i}$ from the LLM, denoted as $\hat{y_i} \subset e_i$. Subsequently, we construct a new dataset $D_{exp} = \{(x_i, e_i)\}^N$, where $1 \le i \le N$. The objective during the training phase is to effectively utilize the generated content and learn from two distinct tasks: the explanation task utilizing data from $D_{exp}$ and the prediction task utilizing the data from the original dataset $D_{pre} = D = \{(x_i, y_i)\}^N$, where $1 \le i \le N$.
To further enhance reasoning performance, we introduce the reasoning verification mechanism within the existing multi-task learning framework. This mechanism enhances the learning process by providing verification signals for additional-task learning. The details will be elaborated in the subsequent sections.

\subsubsection{Learning with Rationale}
To train the proficient reasoners, we employ the multi-task learning framework and divide the learning task into explanation and prediction, where explanation tends to furnish the rationale based on the input sample and the task objective, while the prediction task focuses solely on inferring sentiment polarity. Through the implementation of multi-task learning, the training phase incorporates the losses associated with both explanation and prediction tasks. The loss function is delineated as follows, where $\mathcal L_{exp}$ is the loss for explanation task and $\mathcal L_{pre}$ is the loss for prediction task:
\begin{align}
\mathcal L_{loss} = \alpha \mathcal L_{exp} + (1-\alpha) \mathcal L_{pre}
\end{align}

where the prediction $\mathcal L_{pred}$ aims to minimize the cross-entropy loss for label prediction:
\begin{align}
\mathcal L_{pre} = \frac{1}{N} \sum_{N}^{i=1} \ell_{CE}(\hat{y_i}, y_i)
\end{align}

while the explanation loss $\mathcal L_{exp}$ tends to minimize the generation loss for the rationale, and there exists a subtle distinction between reasoning (RE) and rationalization (RA) scenarios.
\begin{align}
RE: \mathcal L_{exp} = \frac{1}{N} \sum_{N}^{i=1} \ell_{CE}(f(x_i,t_i), \hat{e_i})
\end{align}
\begin{align}
RA: \mathcal L_{exp} = \frac{1}{N} \sum_{N}^{i=1} \ell_{CE}(f(x_i,t_i,y_i), \hat{e_i})
\end{align}
The objective is to equip the model with proficiency in both explanation and prediction, thereby enhancing its reasoning capabilities. However, during the inference phase, only the prediction task is required for evaluation to optimize the inference efficiency and mitigate computational costs.

\subsubsection{Reasoning with Verification}

Considering the rationale generated by LLM is directly employed without any post-filtering processes, it might introduce some error patterns that can negatively influence the performance of multi-task fine-tuning. Some research works perform answer-based filtering to improve the rationale quality. \cite{Explanations} directly removed the incorrect rationale given by reasoning prompting based on the final prediction and supplemented it with the rationale generated under rationalization prompting to complete the final rationale set for training. \cite{HoSY23} demonstrated that answer-based filtering can also be compensated by a diversity of reasoning paths using diverse reasoning and retaining the rationales leading to the correct answer. In our approach, we preserve the sets of rationales generated by the LLM by introducing a verification signal to facilitate further analysis of rationale quality within the multi-task learning framework. This is achieved by incorporating an additional task for verification.

Specifically, we leverage the rationale set generated by our TH-RE prompting and adopt answer-based verification according to the prediction label $\hat{y_i}$ provided by the LLM and the ground truth annotation $y_i$. Rationales that lead to the correct answer label are deemed to possess higher quality and utility compared to those inferring an incorrect answer label. Based on this premise, we complete the prompting using the following template:
\begin{tcolorbox}[colback=gray!10,
                  colframe=black,
                  arc=1mm, auto outer arc,
                 ]
 Given the rationale $e_i$, Please verify whether the above given rationale is reasonable. Return True or False.
\end{tcolorbox}

To generate the verification signal $v_i$, we validate the reasoning rationales that successfully infer the correct answer and provide a general signal as unreasonable with an \emph{False} label for the other rationales, indicating that they could benefit from further refinement. However, according to our observation, LLMs with larger parameter scales, like GPT-3.5-Turbo, tend to predict ambiguous answers containing dual polarities when faced with uncertainty in making a final judgment. Therefore, we establish the verification signal $v_i$ based on the following criteria: 
\begin{align}
    \begin{cases}
      & \text{if } \hat{y_i} \subset \{\hat{y}_{i(t_1)},\hat{y}_{i(t_2)}\},\ \hat{y_i} = \hat{y}_{i(t_1)}, \\
      & \text{if } \hat{y_i} = y_{i},\  v_i \ is \ True
    \end{cases} 
\end{align}

In cases where the rationale presents two polarities, $\hat{y}_{i(t_1)}$ and $\hat{y}_{i(t_2)}$, where $t_2 > t_1$, answer-based verification is conducted on $\hat{y}_{i(t_1)}$ based on the First-Fome-First-Served (FCFS) rule, since the label generated earlier is regarded as holding a greater likelihood according to the next token generation. Then the revised loss function incorporating verification signals is formulated as follows:  
\begin{align}
\mathcal L_{loss} = \alpha \mathcal L_{exp} + \gamma \mathcal L_{ver} + (1-\alpha - \gamma) \mathcal L_{pre}
\end{align}

where the verification loss concerns the self-validation outcome under the supervision of the verification signal:
\begin{align}
\mathcal L_{ver} = \frac{1}{N} \sum_{N}^{i=1} \ell_{CE}(f(e_i), \hat{v_i})
\end{align}

%% file: src/experiment.tex
\section{Experiments}

\subsection{Setups}
In the experiments, we evaluate the results on Restaurant and Laptop datasets in SemEval-2014 \cite{Pontiki_Galanis_Pavlopoulos_Papageorgiou_Androutsopoulos_Manandhar_2014}. To test the performance for ISA, we follow the prior works utilizing datasets that further labeled with explicit and implicit tags \cite{Li2021LearningIS}. To generate effective rationales conducive to reasoning learning, we make use of DO LLMs, Vicuna-13B \cite{zheng2023judging} and GPT-3.5-turbo \cite{CHATGPT} in stage 1 for rationale preparation. Considering the impressive performance of ED style models in understanding input information and comprehension among different tasks, Flan-T5\cite{flant5} serves as the backbone LLM during the multi-task fine-tuning stage. We test with different sizes of Flan-T5, scaling from the base model (250M) to the XXL model (13B). For the baseline methods, we compared with the recently reported best results, including seven baseline methods, which are BERT+SPC \cite{Devlin_Chang_Lee_Toutanova_2019}, BERT+ADA \cite{rietzler2019adapt}, BERT+RGAT \cite{wang-etal-2020-relational}, BERT$_{Asp}$+CEPT\cite{Li2021LearningIS},  BERT$_{Asp}$+SCAPT\cite{Li2021LearningIS}, THOR \cite{Fei_Li_Liu_Bing_Li_Chua} and ABSA-ESA \cite{Ouyang_Yang_Liang_Wang_Wang_Li_2024}. Among them, THOR \cite{Fei_Li_Liu_Bing_Li_Chua} stimulates performance based on CoT prompting with three-step generation. Compared to their method, we utilize a multi-task learning framework during training while directly inferring the final prediction during inference time. To identify the optimal hyperparameters in the training loss, a greedy search is undertaken using the validation set to determine the final values of $\alpha$ and $\gamma$. Without verification supervision, we get the best result when  $\alpha = 0.5$ with explanation and prediction tasks only. With the verification supervision, we get the greatest performance when $\alpha = \gamma = 0.3$. The following experiments will follow this hyperparameter setting.

\begin{table*} 
  \caption{Main results compared with baselines on Restaurant and Laptop datasets. The results with $^\dagger$ and $^\star$ are obtained from \cite{Li2021LearningIS} and \cite{Ouyang_Yang_Liang_Wang_Wang_Li_2024}, while the other results are self-rerun or self-implemented. In our methods, the subscripts stand for learning from rationales generated by different models, which are Vicuna-13B($v$) and GPT-3.5-turbo($g$), respectively. The subscripts $A$ and $F$ represent the accuracy and macro-F1 score.}
  \label{main_result}
  \centering
  \begin{tabular}{lccc|ccc}
    \toprule[1.2pt]
         \makebox[0.25\textwidth][c]{} & \multicolumn{3}{c}{Restaurant} & \multicolumn{3}{c}{Laptop}\\
     \cmidrule(r){2-7}
     & All$_A$ & All$_F$ & ISA$_A$ & All$_A$ & All$_F$ & ISA$_A$ \\
    \hline
    \bf\emph{- State-of-the-art baselines}&&&&&&  \\
    BERT + SPC$^\dagger$ (110M) \cite{Devlin_Chang_Lee_Toutanova_2019} & 83.57 & 77.16 &65.54 & 78.22 & 73.45 & 69.54\\
    BERT + ADA$^\dagger$ (110M) \cite{rietzler2019adapt} & 87.14 & 80.05 &65.92 & 78.96 & 74.18 &70.11 \\
    BERT + RGAT$^\dagger$ (110M) \cite{wang-etal-2020-relational} & 86.60 & 81.35 &67.79 & 78.21 & 74.07 &72.99 \\
    BERT$_{Asp}$ + CEPT$^\dagger$ (110M) \cite{Li2021LearningIS} &87.50 & 82.07 &67.79 & 81.66 & 78.38 &75.86  \\
    BERT$_{Asp}$ + SCAPT$^\dagger$ (110M) \cite{Li2021LearningIS} & 89.11 & 83.79 & 72.28& 82.76 & 79.15 & 77.59\\
    T5$_{Base}$ + ABSA-ESA$^\star$ (220M) \cite{Ouyang_Yang_Liang_Wang_Wang_Li_2024} &88.29&81.74&70.78&82.44&79.34&80.00\\
    \hline
    \bf\emph{- Prompt-based methods}&&&&&& \\
    Flan-T5 + prompt (250M) & 86.88  & 79.78 &  65.17  & 81.98  & 77.93 &  73.71  \\
    Flan-T5 + prompt (11B) &89.29&83.68&75.28 & 81.82 & 77.69 & 75.43\\
    Flan-T5 + THOR (250M) \cite{Fei_Li_Liu_Bing_Li_Chua} &87.68&81.10&68.54&81.66& 77.51&74.29\\
    
    Flan-T5 + THOR (11B) \cite{Fei_Li_Liu_Bing_Li_Chua} &88.57&82.93&73.03&82.29&78.78&76.57\\ 
    \hline
    \bf\emph{- Our methods}&&&&&& \\
    Flan-T5 + RVISA$_v$ (250M) &86.43&78.49&65.92&80.72&76.49&73.71\\
    Flan-T5 + RVISA$_g$ (250M) &86.61&78.92&66.67&81.19&77.13&75.43\\
    Flan-T5 + RVISA$_v$ (11B) &91.25&86.57&81.65&86.52&83.28&87.43\\
    Flan-T5 + RVISA$_g$ (11B) &\bf{91.52}&\bf{86.85}&\bf{82.02}&\bf{86.68}&\bf{84.05}&\bf{88.00}\\
    \bottomrule[1.2pt]
  \end{tabular}
\end{table*}

\begin{table}
  \caption{Results compared with THOR \cite{Fei_Li_Liu_Bing_Li_Chua}. The evaluation metric is the F1 score trained with Flan-T5. The results with $^\dagger$ are self-rerun using the source code from \cite{Fei_Li_Liu_Bing_Li_Chua}.}
  \label{compare_thor}
  \centering
  \begin{tabular}{lllll}
    \toprule
        & \multicolumn{2}{c}{Restaurant} & \multicolumn{2}{c}{Laptop}\\
     \cmidrule(r){2-5}
     & All& ISA& All&ISA\\
     \toprule
     Prompt$^\dagger$ (11B) &83.68 &74.48 & 77.69& 72.44\\
     THOR$^\dagger$ (11B) &82.93&73.08&78.78&72.82\\
     \toprule
      RVISA$_v$ (11B) &86.57&81.73&83.26&85.36\\
    RVISA$_g$ (11B) &\bf{86.85}&\bf{82.61}&\bf{84.05}&\bf{86.20}\\
    \bottomrule
  \end{tabular}
\end{table}
\begin{table}
  \caption{Ablation study of three-hop prompting (TH) and verification (VE) with F1 score metric.}
  \label{ablation}
  \centering
  \begin{tabular}{lllll}
    \toprule
        & \multicolumn{2}{c}{Restaurant} & \multicolumn{2}{c}{Laptop}\\
     \cmidrule(r){2-5}
     & All& ISA& All&ISA\\
     \toprule
      RVISA$_v$ &86.57&81.73&83.26&85.36\\
      - w/o VE &85.91&80.40&82.63&83.39\\
      - w/o VE and TH &85.79&79.10&82.57&82.91\\
      \toprule
    RVISA$_g$ &\bf{86.85}&\bf{82.61}&\bf{84.05}&\bf{86.20}\\
    - w/o VE &86.16&80.32&82.51&83.83\\
    - w/o VE and TH &85.60&79.68&82.05&83.14\\
    \bottomrule
  \end{tabular}
\end{table}
\begin{figure*}
\includegraphics[width=\textwidth]{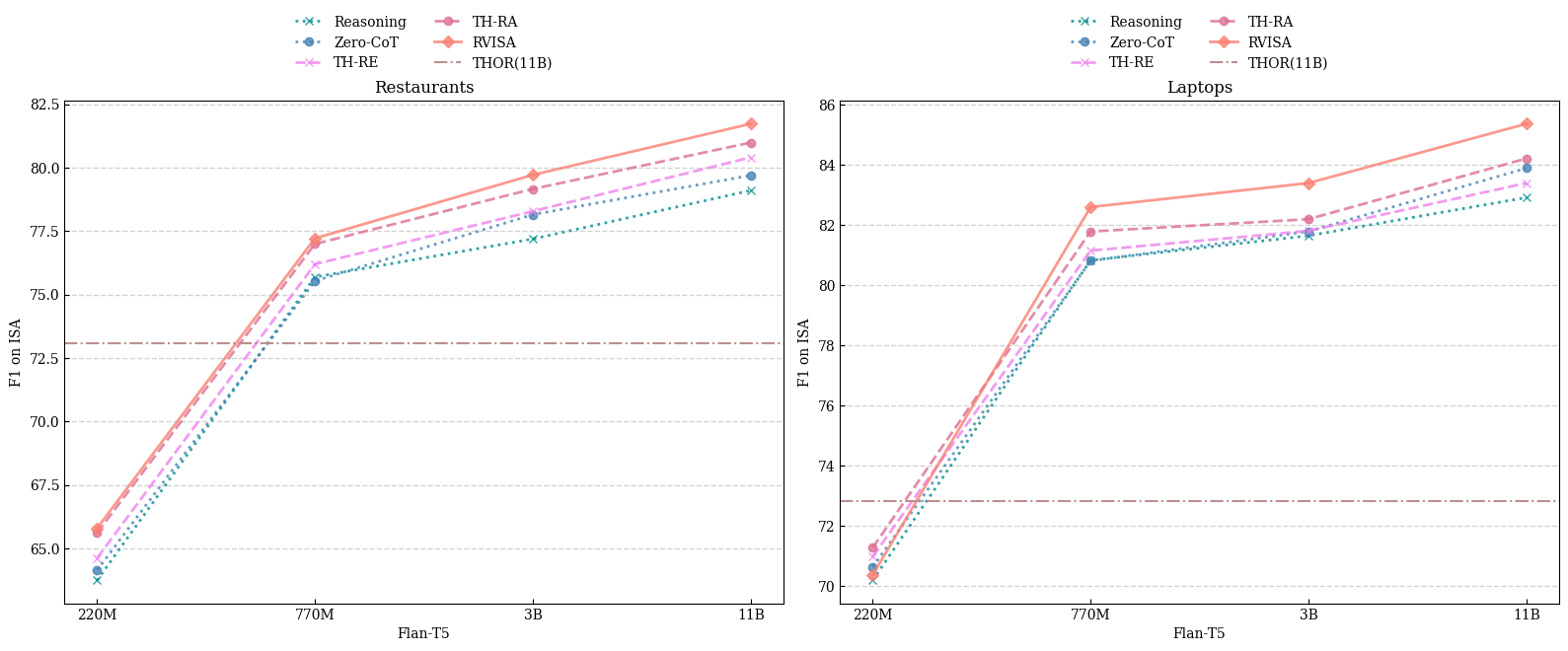}
\caption{The impact of diverse rationales and different model sizes on implicit F1 score. The dashed horizontal line represents the best result of THOR rerun with the Flan-T5-XXL(11B) model on the implicit dataset.}
\label{fig:impact}
\end{figure*}
\subsection{Main Results}
\paragraph{Multi-task learning outperforms the baselines}
The main results of baselines and our method, RVISA, are demonstrated in Table \ref{main_result}. The evaluation metrics include Accuracy and Macro-F1 score. Notably, as THOR \cite{Fei_Li_Liu_Bing_Li_Chua} does not provide the accuracy outcome for implicit sentiment, we rerun the results based on the provided source code. It can be seen that RVISA significantly outperforms the baseline methods, irrespective of whether learning is from Vicuna-13B or GPT-3.5-turbo, underscoring the efficacy of learning within the proposed multi-task learning framework. 

\paragraph{Strong teachers lead to higher quality learning}
The performance of RVISA$_g$ training under the assistance of GPT-3.5-turbo exhibits enhanced reasoning capabilities in implicit sentiment inference compared to RVISA$_v$ trained by using the rationales generated by Vicuna-13B. This disparity can be attributed to the superior common sense knowledge and reasoning prowess exhibited by GPT-3.5-turbo in producing high-quality rationales, which play a pivotal role in transferring reasoning abilities to the Flan-T5 backbone model. However, the smaller backbone model with a base size (250M) lags behind some of the baseline methods due to its limited generation capacity to derive advantage from rich knowledge through in-context learning.

\paragraph{Explicit rationales learning helps implicit reasoning}
We further compare our method with THOR, which is built upon a chain-of-thought strategy. Instead of eliciting the reasoning ability of the language model through sequential three-step prompting, our method stands out by explicitly giving the rationales as informative resources equipped with a verification mechanism to ensure the learning quality. The comparative results are depicted in Table \ref{compare_thor}. RVISA demonstrates superior performance over THOR in terms of F1 score for implicit sentiment analysis while maintaining competitive results in overall F1 score. This underscores the effectiveness of our method in learning implicit sentiment through reasoning tasks and adeptly capturing implicit relationships among instances. Although THOR claimed the three-step generation during fine-tuning can unleash the reasoning power of the backbone model, the re-run result demonstrated limited improvement in F1 scores. This suggests the vulnerability of THOR to enhance prompt-based inference depending on the backbone model (i.e., Flan-T5 \cite{flant5}) itself. In contrast, our method prioritizes effective learning from high-quality sentiment information and closely related tasks, offering a more coherent and justifiable approach to achieving high performance in implicit sentiment analysis.


\subsection{Ablation Study}

We conducted an ablation study on the three-hop prompting (TH) and verification mechanism (VE) components, the results of which are summarized in Table \ref{ablation}. Our analysis compares the F1 scores in both overall and implicit sentiment scenarios. The findings indicate that the absence of the verification mechanism leads to performance degradation in both cases, with a more significant decline of over one point observed in the implicit sentiment results. This highlights the critical role of verification signals in the context of reasoning learning from LLMs, as the answer-based mechanism aids the backbone model in identifying potential errors or unreasonable attributes during multi-task learning processes.

In addition, performance goes down even more when we reduce the CoT prompting from three-hop reasoning to reasoning prompting alone, without the sentiment elements to help guide rationale generation. This happens in both implicit and general scenarios. These observations persist regardless of whether the rationales are generated by Vicuna-13B or GPT-3.5-turbo, indicating that, irrespective of generation quality, the three-hop prompting mechanism plays a pivotal role in steering the correct direction of reasoning for implicit sentiment analysis. Although the impact of performance degradation with three-hop reasoning prompting is less pronounced compared to the absence of the verification mechanism, it is evident that their contributions are mutually reinforcing and indispensable. It is also the essence of multi-task learning, where tasks are strongly related and complement each other.

\subsection{Further Analysis}

\paragraph{The impact of rationale}
In our investigation of the influence of diverse rationales, we conducted training experiments using rationales generated by Vicuna-13B with various prompting methods, including Reasoning, Zero-CoT, Three-hop Reasoning (TH-RE), and Three-hop Rationalization (TH-RA), as depicted in Figure \ref{fig:rationale}. We compared the results with RVISA and THOR, as shown in Figure \ref{fig:impact}, where RVISA is enhanced by Three-hop Reasoning prompting with the verification mechanism. It can be seen that the model trained with TH-RA demonstrates the second-best results since rationalization prompting can leverage the gold answer as context information to elucidate the underlying logic. This approach facilitates the generation of more reasonable rationales that lead to correct answers. Consequently, TH-RA generally outperforms TH-RE, where TH-RE may produce more problematic responses, resulting in incorrect answers. However, RVISA consistently outperforms both TH-RA and TH-RE, suggesting that the language model, when trained under verification signals, can leverage erroneous or irrational attributes present in TH-RE-generated rationales. This provides a visible solution to utilize LLM-generated labels as an additional verification factor. Furthermore, rationales generated by Reasoning and Zero-CoT methods lag behind Three-hop prompting in most scenarios, underscoring the importance of our designed prompting approach in structuring coherent rationales and eliciting highly relevant sentiment elements within the three-hop prompting.

\paragraph{The impact of model size}

Figure \ref{fig:impact} also illustrates the impact of backbone model size on reasoning learning. In the Restaurant dataset, smaller-sized models (i.e., base and large) exhibit marginal performance improvements under the verification mechanism, indicating the limited capabilities of small models to benefit from the prompt-based inference within the multi-task framework. However, as model size increases, the combined benefits of the verification mechanism and three-hop reasoning prompting demonstrate enhanced potential, leading to a widening performance gap compared to the second-best TH-RA method. Notably, with large (770M) size models, RVISA achieves superior performance to the best result of THOR trained with the Flan-T5-XXL (11B) model on both Restaurant and Laptop datasets, showcasing the efficacy of our method in enhancing reasoning abilities for pre-trained models. When it comes to the XXL (11B) size, TH-RE, TH-RA, and RVISA collectively surpass THOR in implicit sentiment prediction. The Laptop dataset demonstrates similar trends. All prompting methods with XXL size model under the multi-task learning framework surpass the best result of THOR, emphasizing the effectiveness of our proposed framework and the scaling effect influenced by the learning capabilities of the trained model. 


%% file: src/discussion.tex
\section{Discussion}
We propose a two-stage reasoning framework, RVISA, to learn effectively and reliably from the rationales generated by DO LLMs for implicit sentiment analysis. We show that RVISA holds the potential to promote the reasoning and learning ability of ED model under the supervision of verification through extensive experiments. In this section, we discuss the error scenario after fine-tuning using our proposed method and the limitations for further improvements.

\begin{figure}[h]
\includegraphics[width=0.48\textwidth]{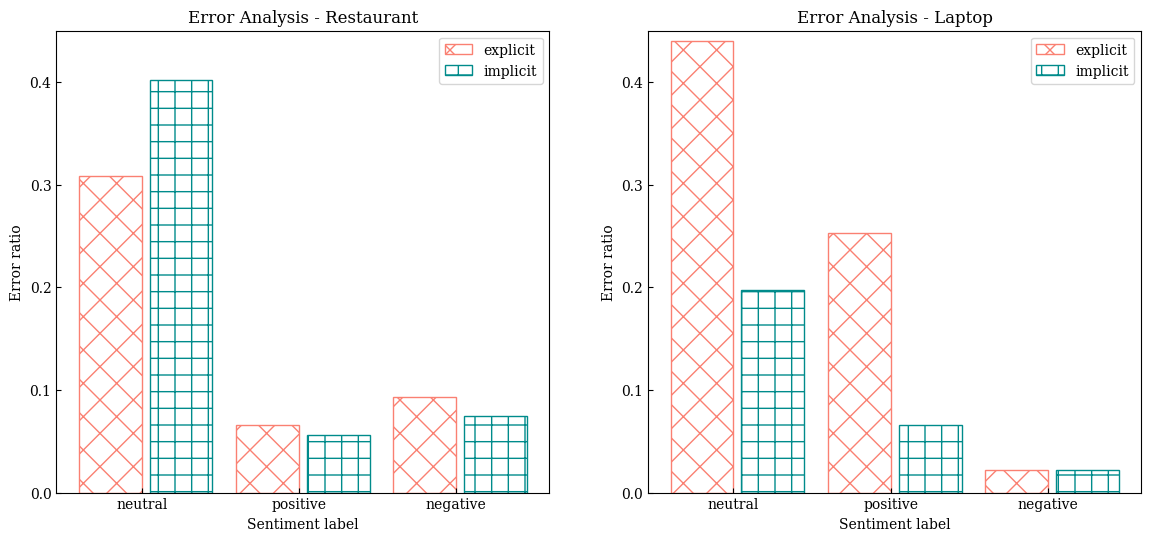}
\caption{Error analysis for two datasets with rationales generated by GPT-3.5-turbo. The error ratio here refers to the proportion of the number of error types to the total number of error instances.}
\label{fig:error}
\end{figure}

\paragraph{Error Analysis} Our proposed method demonstrates superior performance in implicit sentiment analysis. To further explore the error scene, we calculate the error ratio considering sentiment types, including explicit and implicit, and with the relationship to the corresponding sentiment labels. The result is shown in Figure \ref{fig:error} with the rationales generated from GPT-3.5-turbo.  It can be observed that for the Laptop dataset, errors in neutral predictions within the explicit dataset surpass those in the implicit dataset, resulting in the F1 score performance in the implicit dataset exceeding that in all data. Conversely, in the Restaurant dataset, the error ratio associated with neutral predictions in the implicit dataset exceeds that of the original dataset. This observation underscores the significant influence of neutral sentiment distribution on error distribution patterns. Moreover, the ratio of incorrect predictions pertaining to \emph{neutral} polarity exceeds $60\%$. This suggests the nuanced challenges associated with accurately discerning neutral sentiments within sentiment analysis tasks, highlighting the need for further refinement and optimization in model training and inference processes.

\paragraph{Limitations} In this study, we propose a straightforward yet effective verification mechanism to enhance the overall performance in sentiment analysis. The answer-based verification plays a key role in the RVISA framework, demonstrating its significance in reasoning learning. It is worth noting that while the current answer-based verification signal is effective, there is potential for further enhancement through the exploration of alternative verification modes or the incorporation of additional pertinent factors. This avenue for future research paves the way for more nuanced and reliable sentiment analysis. On the other hand, the three-hop prompting proves instrumental in generating effective rationales by deducing sentiment elements. It is manually designed with the format drawing on prior works, which poses challenges in further optimization. Given the evolving landscape of advanced techniques focused on optimizing prompts for LLMs, it is unclear whether the prompt can be generated automatically or optimized through the utilization of soft prompts in this study. This raises a feasible direction for further exploration.

%% file: src/conclusion.tex
\section{Conclusions}

In conclusion, this study sheds light on implicit sentiment analysis in the era of LLMs and proposes a novel two-stage learning framework, RVISA, designed to incorporate reasoning and verification for implicit sentiment analysis. By leveraging the generative prowess of DO LLMs, we empower ED backbone models with enhanced reasoning capabilities. The utilization of three-hop reasoning prompting facilitates the explicit generation of cues guided by sentiment element construction, which is conducive to reasoning learning. Through a straightforward and effective answer-based verification mechanism, we ensure robust and reliable reasoning learning to further improve the proficiency of our ED backbone model in inferring implicit sentiment. The experimental results demonstrate superior performance and achieve state-of-the-art results in ISA on two benchmark datasets.

%% file: src/Appendix.tex
\section*{Rationale Generation}
\paragraph{Vicuna-13B versus GPT-3.5-turbo}

\begin{figure}[h!t]
\includegraphics[width=0.48\textwidth]{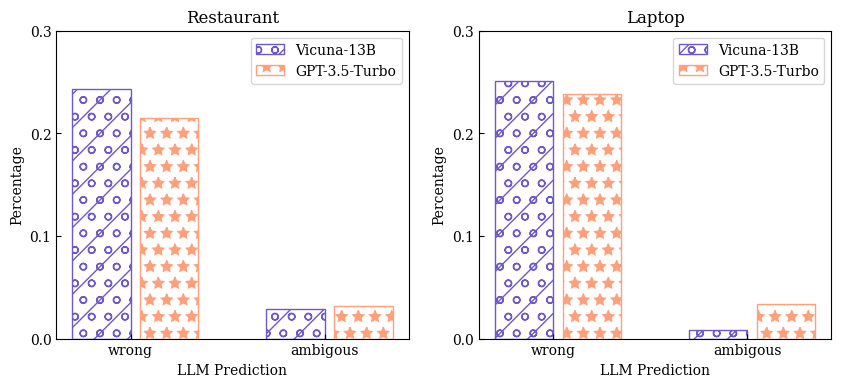}
\caption{Wrong prediction and ambiguous prediction analysis for Vicuna-13B and GPT-3.5-turbo, where ambiguous stands for multiple uncertain predictions in the rationale.}
\label{fig:ambigous}
\end{figure}

To delve into the quality of rationales generated from Vicuna-13B and GPT-3.5-turbo, the analysis for wrong and ambiguous prediction is conducted as illustrated in Figure \ref{fig:ambigous}. In both Restaurant and Laptop datasets, Vicuna-13B exhibited a slightly higher count of incorrect predictions compared to GPT-3.5-turbo. This suggests that stronger models such as GPT-3.5-turbo demonstrate a superior capability to generate higher-quality rationales, leading to more accurate final predictions. However, the percentage of ambiguous predictions originating from GPT-3.5-turbo surpassed that of Vicuna-13B, which indicates that the more powerful model exhibits a greater tendency to generate uncertain expressions rather than provide definitive judgments when deciphering the nuanced sentiment. It also underscores the inherent challenge of capturing subtle nuances in sentiment within constrained contextual information.

\paragraph{Reasoning (Re) versus Three-hop Reasoning (TH-RE)}

Table \ref{re_vs_thre} shows the results when generating the rationale with GPT-3.5-turbo using diverse prompting. Under the reasoning prompt, the LLM tends to break down the problem into four fine-grained steps, leveraging its own pre-trained knowledge. However, even with more steps of inference, the prediction is the complete opposite of the gold label. On the other hand, with the guidance of three-hop prompting, the rationale follows the generation format as shown in the prompting template, which is more concise and structured to infer the sentiment elements leading to the final correct prediction. 
\begin{table}[h!t]
  \caption{Reasoning(Re) prompting compared with Three-hop Reasoning(TH-RE) prompting for GPT-3.5-turbo model.}
  \label{re_vs_thre}
  \centering
  \begin{tabular}{lp{6cm}}
    \toprule
     \textbf{Input Text} & a cheaper price should not equal a ``cheap'' product.\\
     \textbf{Aspect Term} & price \\
     \textbf{Sentiment Polarity} & positive \\
     \toprule
     \textbf{Re Prompt} & Given the sentence \{input text\}, what's the sentiment polarity towards price, why? \\
     \toprule
     \textbf{GPT-3.5-turbo} & The sentiment polarity towards price in this sentence is {\color{red}negative}. The sentence suggests that a lower price does not necessarily mean that the product is of low quality or ``cheap.'' In other words, the sentence implies that there is a common misconception that cheaper products are of inferior quality, which is not necessarily true. To break it down step by step: 1. The sentence contains the word ``cheaper,'' which is a synonym for ``lower-priced'' or ``more affordable.'' This suggests that the speaker is concerned with the price of the product. 2. However, the sentence also contains the word ``cheap,'' which is often used to describe products of low quality or value. This suggests that the speaker is concerned with the perception of the product's quality. 3. The sentence negates the idea that a cheaper price automatically equals a ``cheap'' product. This suggests that the speaker believes that there is a common misconception that cheaper products are of inferior quality, which is not necessarily true. 4. Therefore, the sentiment towards price is negative because the sentence implies that people often assume that cheaper products are of low quality, which is not always the case. \\
     \toprule
     \textbf{TH-RE Prompt} & Given the sentence \{input text\}, what's the sentiment polarity towards price, why? Let’s think step by step. \textbf{The mentioned aspect} towards price is about ... \textbf{The underlying opinion} towards price is about ... \textbf{Therefore, the sentiment polarity} towards price is ...\\
     \toprule
     \textbf{GPT-3.5-turbo} & \textbf{The mentioned aspect} towards price is about the relationship between price and the quality of a product. \textbf{The underlying opinion} towards price is that just because something is offered at a lower price, it should not automatically be assumed to be of low quality or ``cheap.'' \textbf{Therefore, the sentiment polarity} towards price in this sentence is {\color{red}positive}. The sentiment is emphasizing that price should not be the sole determinant of product quality, advocating for a more nuanced understanding of the pricing and value relationship.\\
     
    \bottomrule
  \end{tabular}
\end{table}